\PassOptionsToPackage{numbers,square}{natbib} 
\documentclass{article}
\usepackage[final]{neurips_2019}
\usepackage[utf8]{inputenc} 
\usepackage[T1]{fontenc}    
\usepackage{hyperref}       
\usepackage{url}            
\usepackage{booktabs}       
\usepackage{amsfonts}       
\usepackage{nicefrac}       
\usepackage{microtype}      
\usepackage{tikz, makecell, algorithm, algpseudocode}
\usepackage{tikz}
\usetikzlibrary{arrows,positioning,shapes}
\usepackage{booktabs}

\title{Preserving Patient Privacy while Training a Predictive Model of In-hospital Mortality}

\author{%
  Pulkit Sharma, \hspace{0.1cm} Farah E Shamout, \hspace{0.1cm} David A Clifton
  \\
  Department of Engineering Science\\
  University of Oxford\\
  \texttt{\{pulkit.sharma, farah.shamout, david.clifton\}@eng.ox.ac.uk} \\
}

\begin{document}

\maketitle

\begin{abstract}
Machine learning models can be used for pattern recognition in medical data in order to improve patient outcomes, such as the prediction of in-hospital mortality. Deep learning models, in particular, require large amounts of data for model training. However, the data is often collected at different hospitals and sharing is restricted due to patient privacy concerns. In this paper, we aimed to demonstrate the potential of distributed training in achieving state-of-the-art performance while maintaining data privacy. Our results show that training the model in the federated learning framework leads to comparable performance to the traditional centralised setting. We also suggest several considerations for the success of such frameworks in future work. 
\end{abstract}

\section{Introduction}
In recent years, deep learning has achieved state-of-the-art performance that surpasses human level performance across various tasks in computer vision, speech recognition, and natural language processing \cite{Karen_VGG_14,hinton_sr_shared_views_SPM_12,Jacob_BERT_18}. One of the key drivers of this success is the availability of large amounts of data to train the deep learning models. In applications where privacy is not a concern, datasets are usually curated in a central location and may even be publicly available for further research \cite{awesomedata_2019}.

Although we are currently in the era of data-rich medicine, clinical data often exists in isolation due to several privacy reasons. Some of those issues include potential invasion of privacy, data misuse, or patient discrimination. Basic data anonymisation techniques, such as those listed in the guide published by the Personal Data Protection Commission in Singapore \cite{Singapore2018}, are also at risk of de-anonymisation through reverse engineering. Due to the sensitivity of medical data, most existing clinical databases are curated for private use only \cite{Shickel2017}, and only a few are publicly available \cite{MIMIC3_16}. This hinders the progress of developing deep learning frameworks using diverse datasets in order to improve patient outcomes. 

In order to address such concerns, various governments have made initiatives to strengthen data privacy and security \cite{vito_fed_19}, such as the General Data Protection Regulation (GDPR) that was enforced in 2018 by the European Union. Privacy is also a key value of the Montreal Declaration for a Responsible Development of Artificial Intelligence (2018) \cite{montrealdec}. These regulations make the use of traditional centralised machine learning models more challenging, where the data is collected from multiple parties and then processed on a central server.

In this paper, we test a privacy preserving framework for the task of in-hospital mortality prediction amongst patients admitted to the intensive care unit (ICU), as in related works \cite{shamout2019deep}. We use federated learning (FL) \cite{McMahan_AISTATS17,Jakub_FL_15,Jakub_FL_16}, which involves training a global machine learning model using vital-signs data distributed across remote devices (e.g. in various hospitals), without having to share the data with a centralised server. The FL framework provides a solution for all the issues related to privacy, locality and ownership of healthcare data \cite{Keith_FL_19}. Based on the results, we discuss the strengths and challenges that are unique to the potential of FL within healthcare and offer recommendations for relevant policy-making parties. 






\section{Problem Formulation} 
\label{sec:probform}
\begin{figure*}[!ht]
	\centering
	\tikzstyle{blockVS} = [rectangle, draw, line width=1.2pt, fill=blue!10,
	text width=1.25cm, text centered, rounded corners, minimum height=2.2em]
	\tikzstyle{blockRS} = [diamond, draw, line width=1.2pt, fill=yellow!10,
	text width=1.1cm, text centered, rounded corners, minimum height=2.2em]
	\tikzstyle{line} = [draw, line width=1.2pt, -latex']
	\tikzstyle{lineD} = [draw, line width=1.2pt, dashed']
	\begin{tikzpicture}[node distance = 2cm, auto]
	\node [] (train_data) {};
	\node [blockVS] (device1) {\bf{Hospital 1}};
	\node [blockVS, right of=device1, node distance=2.27cm] (device2) {\bf{Hospital 2}};
	\node [blockVS, right of=device2, node distance=2.27cm] (device3) {\bf{Hospital 3}};
	\node [blockVS, right of=device3, node distance=4.27cm] (deviceK-1) {\bf{Hospital K-1}};
	\node [blockVS, right of=deviceK-1, node distance=2.27cm] (deviceK) {\bf{Hospital K}};
	\node [above of=device3, node distance=3cm] (temp1) {};
	\node [blockRS, right of=temp1, node distance=0.7cm] (Server) {\bf{Common Server}};
	\path [line, dotted] (device1) -- (Server) node[pos=0.5,right] {\makecell[l]{\textbf{Local}\\ \textbf{Updates}}};
	\path [line, dotted] (device2) -- (Server);
	\path [line, dotted] (deviceK) -- (Server) node[pos=0.5,right] {\makecell[l]{\textbf{Local}\\ \textbf{Updates}}};
	\path [line, dash dot] (Server) -- (device3) node[pos=0.5,right]
    {\makecell[l]{\textbf{New Global}\\ \textbf{Model}}};
	\path [line, dash dot] (Server) -- (deviceK-1);
	\end{tikzpicture}
	\caption{Schematic of the federated learning (FL) framework adopted for the in-hospital mortality prediction task. In order to preserve the privacy of clinical data, the model is trained in a distributed fashion: The hospitals  periodically communicate  the local updates with a common server to learn a global model. The common server incorporates the updates and sends back the parameters of the updated global model.}
	\label{fig:intro_block_dia}
\end{figure*}
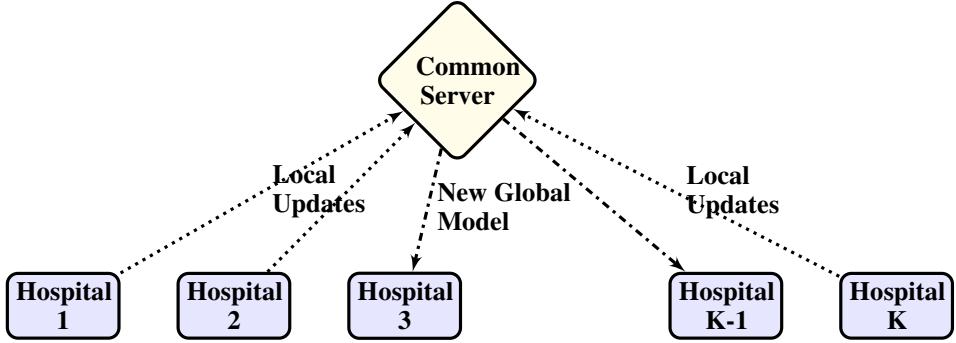

We aim to securely train a deep learning model, referred to thereafter as the global model, that can predict in-hospital mortality for ICU patients. The mortality prediction task is formulated as a binary classification problem, where the label indicates patient's death before hospital discharge. The block diagram depicting the proposed approach is shown in Figure \ref{fig:intro_block_dia}. 
The data stored locally at different hospitals is used to estimate the local (model) updates which are communicated  to a common server without sending the data. 
The common server then employs these updates to derive a better global model which is further used during testing at different hospitals.

To perform the secured model training, we assume that a set of hospitals $\mathcal{H} = \{\mathcal{H}_1, \dots, \mathcal{H}_K\}$ participate in training a global machine learning model (also referred to as the training federation) with a common server $S$ coordinating between them. Each hospital $\mathcal{H}_k$ stores its data $D_k = \{(x_{1}^{k}, y_{1}^{k}), (x_{2}^{k}, y_{2}^{k}), \dots  (x_{|D_k|}^{k}, y_{|D_k|}^{k})\}$ locally and does not share it with $S$. 
In this work, $x_{i}^{k}$ and $y_{i}^{k}$ represent the data sample $i$ and its corresponding label, respectively,  at hospital $k$. $|D_k|$ represents the total number of data samples stored at hospital $k$. Those assumptions make the FL a suitable choice for machine learning training using clinical data in hospitals where data security and privacy are of utmost importance.  


The training procedure relies on efficient communication between the central server $S$ and the distributed hospitals. The objective is to minimise the global loss function to estimate the global ($G$) parameters $\mathbf{w}^G\in\mathbb{R}^d$ of the global model without directly accessing the data stored at the hospitals, where $d$ represents the number of parameters of the model. 

The common server $S$ first broadcasts the global model $\mathbf{w}_{t}^{G}$ to a subset of non-identically-distributed hospitals $\mathcal{H}^{-}\subset \mathcal{H}$ at time $t$. A local loss is then optimised over the local data $D_k$ at every node $k$ in $\mathcal{H}^{-}$ to estimate the local parameter vector $\mathbf{w}_{t+1}^{k}$. The various hospitals $\mathcal{H}^{-}$ then send their computed model parameters to the common server $S$, which aggregates the findings to estimate $\mathbf{w}^G$ of the final model. This consists of a weighted mean of all the local models to obtain an updated global model $\mathbf{w}_{t+1}^{G}$ as:
\begin{equation}
    \mathbf{w}_{t+1}^{G} = \sum_{k=1}^{|\mathcal{H}^{-}|} p_{t+1}^{k} \mathbf{w}_{t+1}^{k}.
    \label{eq:1}
\end{equation}




For the sake of simplicity, we drop the time dimension and consider only one time instance as: 
\begin{equation}
    \mathbf{w}^{G} = \sum_{k=1}^{|\mathcal{H}^{-}|} p^{k} \mathbf{w}^{k}, 
    \label{eq:2}
\end{equation}
where $p^{k} \in [0,1]$ represents the weights associated with each hospital $k$ such that $\sum_{k=1}^{|\mathcal{H}^{-}|} p^{k} = 1$. The term $p^k$ specifies the relative impact of each hospital and we adopt a general choice of $p^k = \frac{|D_k|}{D}$, where $D = \sum_{k} |D_k|$ is the total number of samples across the $k$ hospitals.
It is worth noting that incorporating the information about data at different hospitals in the hospital-dependent value $p^k$ is important for computing an efficient federated model. 

We iterate through the described training procedure across different hospitals until convergence or some stopping criterion.
Algorithm \ref{table:prpsd_algo} describes the proposed FL setup, where the goal is to estimate model parameters at time $t+1$; i.e. $\mathbf{w}_{t+1}^{G}$ given $\mathbf{w}_{t}^{G}$. 
It has to be noted that at each step the model can be updated locally at each hospital in $k \in \mathcal{H}^{-}$. 
However this model is evaluated using the test data at all the hospitals; i.e. $k \in \mathcal{H}$. 
Accuracy $a_t$ (on test data $k \in \mathcal{H}$) can be used as metric of evaluation while updating the global model where a model is updated if and only if $a_{t+1} \geq a_{t}$.


\begin{algorithm}[t!]
 \caption{ \small A summary of the FL framework to compute the global model at common server using data stored locally at different hospitals. Functions ModelUpdate and LocalTestAccuracy are executed locally on the $k^{th}$ hospital. Variable $a_{t}$ is an estimation of the global accuracy at time $t$.}
 \label{table:prpsd_algo}
 \centering \small
 \noindent
 \begin{flushleft}
\textbf{Input:}  $\mathbf{w}_{t}^{G} \; a_{t}$ \\
 \textbf{Output:} $\mathbf{w}_{t+1}^{G} \; a_{t+1}$   \\
 \end{flushleft}
 \begin{algorithmic}[1] 
 
 \State  broadcast $\mathbf{w}_{t}^{G}$ to hospitals in $\mathcal{H}^{-}$
 \State \textbf{for} each hospital $k \in \mathcal{H}^{-}$ do
 \State \hspace{2em} $\mathbf{w}_{t+1}^{k}    \leftarrow  ModelUpdate(k, \mathbf{w}_{t}^{G})$  
  \State \hspace{2em} $\mathbf{p}_{t+1}^{k}    \leftarrow  \frac{|D_k|}{\sum_{k}|D_k|}$  
  \State \textbf{end for}
  \State  $\mathbf{\tilde{w}}_{t+1}^{G}   \leftarrow \sum_{k=1}^{|\mathcal{H}^{-}|} p_{t+1}^{k} \mathbf{w}_{t+1}^{k}$
  
 \State \textbf{for} each hospital $k \in \mathcal{H}$ do
 \State \hspace{2em} $a_{t+1}^{k}    \leftarrow  LocalTestAccuracy(k, \mathbf{\tilde{w}}_{t+1}^{G})$  
  \State \textbf{end for}
  \State  $a_{t+1}   \leftarrow weighted \; average \;  of \; a_{t+1}^{k} \; \forall \; k \in \mathcal{H}$ 
  \State \textbf{while} $a_{t+1} < a_{t}$
  \State \hspace{2em} $\mathbf{\tilde{w}}_{t+1}^{G}    \leftarrow \mathbf{w}_{t}^{G}$ 
  \State \hspace{2em} $a_{t+1}    \leftarrow a_{t}$ 
  \State \textbf{end while} 
  \State $\mathbf{w}_{t+1}^{G}    \leftarrow \mathbf{\tilde{w}}_{t+1}^{G}$

 \end{algorithmic}
 \end{algorithm}

\section{Experimentation}
\subsection{Dataset}
The proposed FL framework is evaluated for the task of predicting in-hospital mortality using data obtained from the publicly available MIMIC-III database \cite{MIMIC3_16}. The total number of patient admissions for the benchmark mortality prediction task were 21,138, where the variables collected in the first 48-hour window were used as input features \cite{Hrayr_ML_17}. The variables included vital-sign data, such as heart rate and temperature, and details can be found in \cite{MIMIC3_16}. The number of time stamped observations in first 48 hours varied per patient episode. Hence, we used hand-engineered features as described in \cite{Lipton_ICLR_16}. For each vital sign, six different sample statistic features were computed on seven time-series: maximum, minimum, mean, standard deviation, skew and number of measurements. The seven time-series included the full time-series, the first/last 10\% of time, first/last 25\% of time, first/last 50\% of time. The features were then scaled into the range [-1,1] before feeding them into the classifier.

To mimic the FL framework described in Section 2, we distributed the training and testing data amongst virtual workers using the coMind federate learning toolkit \cite{comindorg_2019}. This virtual set up mirrored the realistic scenario such that the data is located physically across different hospitals. 

\subsection{Results}
The primary metrics that we used for evaluation are the area under the receiver operator characteristic curve (AUROC) and area under the precision-recall curve (AUPRC). 
Logistic regression (LR) and a multi-layer perceptron (MLP) classifier were employed in the traditional training procedure such that the training data exists at a central server, denoted as ORG, and in the FL setup. 

The architecture of MLP here was modelled based on \cite{Anna_BNN_19} which consists of a single layer with 50 nodes between the input and output. The networks were trained with a cross-entropy loss and Adam optimiser \cite{Goodfellow-et-al-2016} using a batch size of 8 for 100 epochs. 

The FL setup was simulated with a common virtual server and two local workers (clients/hospitals) with around 1,700 rounds of communication before deriving the final model. The train-test split was the same as in \cite{Hrayr_ML_17} and we split both the train and test data equally among the two local workers. 

The results in form of AUROC and AUPRC are shown in Table \ref{table:FLresults}, where LR-ORG/MLP-ORG and LR-FL/MLP-FL represent the classifiers trained in the traditional and FL setups. 
It can be observed that there is a slight decrease in performance in the case of the FL setup, as opposed to the traditional setting.  

\begin{table}[t!]
\centering  
\caption{Comparison of the proposed FL methods with the standard setup. LR-ORG/MLP-ORG and LR-FL/MLP-FL represents logistic regression/multi-layer perceptron classifier trained in normal and FL setup.}
\label{table:FLresults}
\begin{tabular}{@{}lcccc@{}}
\toprule
               & LR-ORG & LR-FL & MLP-ORG & MLP-FL \\ \midrule
\textbf{AUROC} & 0.8152          & 0.7890         & 0.7925           & 0.7769          \\
\textbf{AUPRC} & 0.4030          & 0.3659         & 0.3900           & 0.3504          \\ \bottomrule
\end{tabular}
\end{table}

\section{Summary and Future Directions}
The performance of the models trained with FL is comparable to training with centralised data. This demonstrates the potential of distributed learning in training machine learning models for clinical tasks. The decline in performance needs further investigation, which could involve further experimentation with the hyper-parameters. While there may be a potential trade-off between privacy and performance, it is evident that FL-based models can generalise for other outcome prediction tasks in healthcare, since it performs well for the in-hospital mortality prediction task. 


Improving data privacy can have a trickle-down effect on other ethical related issues, such as fairness and diversity. With improved data privacy, data owners would be more comfortable in utilising their data for machine learning research by not sharing the data directly. As machine learning models are trained using larger and potentially more diverse datasets, the performance of the models would also improve. Better performing models can then be deployed in clinical trials and eventually improve patient outcomes.

Achieving such long term benefits requires the participation of various stakeholders. First, research efforts must focus on mitigating the limitations of FL and related frameworks. For example, the inaccessibility of the data by the central server compromises model interpretability, which highly relies on examining the original inputs and their respective outputs. This requires further discussion between the parties involved with model development; i.e. the researchers and participating data owners, such as hospitals.  Secondly, the tested framework requires setting up an effective infrastructure across hospitals and research institutions. This infrastructure depends on several resources that must be allocated by involved parties, which include communication, regulation, and funding. 

Privacy-preserving models are vital for the development of machine learning research in the healthcare domain. Future work should focus on improving the performance of those frameworks. Improved performance is also a key consequence of training robust models using diversified and large datasets, while protecting the privacy of the patient - the most important stakeholder.

\bibliographystyle{unsrt}
\bibliography{bib_file}
\end{document}